\documentclass[11pt]{article}
\usepackage[final]{acl}
\usepackage{times}
\usepackage{latexsym}
\usepackage{fontspec}
\usepackage{multirow}
\newfontfamily\devanagarifont[
  Script=Devanagari,
  BoldFont=Shobhika-Bold.otf]{Shobhika-Regular.otf}
\newcommand{\dev}[1]{{\devanagarifont #1}}
\usepackage{microtype}
\usepackage{inconsolata}
\usepackage{float}
\usepackage{amsmath}
\usepackage{graphicx}
\usepackage{caption}
\usepackage{booktabs}
\usepackage{hyperref}
\newcommand\blfootnote[1]{%
  \begingroup
  \renewcommand\thefootnote{}\footnote{#1}%
  \addtocounter{footnote}{-1}%
  \endgroup
}

\title{Seeds Before Objectives: Rethinking Evaluation for Low-Resource Garhwali ASR}

\author{
Karamvir Singh Batra$^{1}$,
Prathamjyot Singh$^{1}$,
Ashima Sood$^{2}$, \\
\textbf{Jasmeet Singh$^{1,*}$,
Sahil Sharma$^{3,*}$} \\
$^{1}$Thapar Institute of Engineering and Technology, Patiala, Punjab, India \\
$^{2}$Ulster University, Londonderry, United Kingdom \\
$^{3}$Ulster University, Belfast, United Kingdom \\
\texttt{\{kbatra\_be22, psingh1\_be22, jasmeet.singh\}@thapar.edu} \\
\texttt{\{sood-a1, s.sharma\}@ulster.ac.uk}
}

\begin{document}
\maketitle
\blfootnote{\raggedright $^{*}$Corresponding authors: \mbox{\texttt{jasmeet.singh@thapar.edu}}, \mbox{\texttt{s.sharma@ulster.ac.uk}}\par}

\begin{abstract}
At corpus sizes typical of low-resource dialects, single-run comparisons can yield gains that do not replicate. We show this for Garhwali, an under-resourced Indo-Aryan language of the central Himalaya, building the first reproducible multi-seed ASR benchmark on the official VAANI splits, with per-seed outputs and significance testing. Re-examining plausible gains, we find them fragile: neither Focal CTC nor a matra-weighted objective beats standard CTC under seed-level testing, the matra objective fails to cut even its targeted errors, and Hindi-to-Garhwali transfer gives no gain over direct fine-tuning. What holds up is mundane: w2v-BERT 2.0 with standard CTC reaches 47.0\% WER over five seeds, beating the larger MMS-1B and comparable models; pretraining design, not parameter count, drives performance, and speed augmentation gives a small, largely consistent gain. Multi-seed evaluation on official splits separates real gains from seed noise.
\end{abstract}

\section{Introduction}
\label{sec:intro}

At the corpus sizes typical of low-resource dialects, comparing systems from single training runs is treacherous: per-utterance error variance is high, the choice of data split materially affects the reported number, and a single lucky seed can manufacture an apparent improvement that disappears on replication. This is not a Garhwali-specific worry but a general methodological one, echoing a broader concern that single-seed results can be unreliable and that benchmark variance must be reported explicitly \citep{bouthillier2021accounting, picard2021torch, bui2025assessing}. Yet low-resource dialectal ASR, where corpora are smallest and the temptation to report a single strong run is greatest, is precisely where variance-aware evaluation is least often practised. \emph{Our central thesis is that, in this regime, multi-seed evaluation on fixed official splits with significance testing is necessary to separate real gains from seed noise and that gains reported from single runs should be treated with caution until replicated.}

We make this case concretely for Garhwali, an Indo-Aryan language spoken by roughly 2.5 million people in the Garhwal Himalaya of Uttarakhand, India. Despite this sizeable speaker base it remains severely under-resourced for speech technology, with little transcribed audio and, until recently, no public ASR benchmark. Its phonology: distinctive matra (vowel-diacritic) patterns, aspirated and retroflex contrasts that diverge from standard Hindi and further complicates recognition. Large multilingual self-supervised speech models have made low-resource ASR tractable through transfer from large-scale pretraining (Section~\ref{sec:related}), raising a natural question for any new dialect: which choices actually matter such as model scale, pretraining coverage, the training objective, or data augmentation? A focal CTC objective, which down-weights easy examples to emphasise hard ones, is a natural candidate for the difficulty imbalance of low-resource dialectal speech—exactly the kind of plausible single-run gain our methodology is designed to test.

We therefore revisit Garhwali ASR with a deliberately rigorous, multi-seed methodology on the \emph{official} VAANI splits, and ask which reported gains are robust. Fine-tuning w2v-BERT 2.0 with standard CTC over five seeds reaches a WER of 47.0\%, outperforming a much larger MMS-1B and comparable-scale models (XLS-R, HuBERT). We also include Whisper Large-v3 as a reference point; we defer its setup and instability to Section~\ref{sec:analysis}. We then test three interventions and find that none reliably helps: neither Focal CTC nor a phonologically-motivated matra-weighted CTC improves over standard CTC once results are averaged over seeds and corrected for multiple comparisons (the matra-weighted objective gains nothing on its targeted error category relative to standard CTC), and a two-stage Hindi$\rightarrow$Garhwali transfer gives no gain over direct fine-tuning. By contrast, speed augmentation yields a small mean gain across all objectives, largely consistent at the seed level. The practical message is that for low-resource dialectal ASR, \emph{pretraining design and data augmentation are reliable levers, whereas objective engineering and cross-lingual transfer are not}.

Concretely, our contributions are:
\begin{itemize}
\item \textbf{A methodological argument, with evidence}, that single-seed comparisons are unreliable at low-resource-dialect corpus sizes: under a multi-seed paired protocol, we re-evaluate three interventions plausible for this setting (Focal CTC, a matra-weighted objective, and Hindi$\rightarrow$Garhwali transfer) and show that none yields a consistent paired advantage over standard CTC; the only seed-consistent gap (standard vs.\ focal) favours the baseline, as the power analysis confirms. Speed augmentation is the only intervention that improves mean WER across all objectives.
\item \textbf{The first reproducible, multi-seed Garhwali ASR benchmark} on the official VAANI splits, with means, standard deviations, bootstrap intervals, and Holm-corrected significance tests over five seeds bringing to a single dialect the transparent, variance-aware evaluation now advocated for ASR more broadly \citep{srivastav2025open}. Code and per-seed results are available at GitHub repository\footnote{\url{https://github.com/soodashima91/Garhwali-ASR}}.
\item \textbf{Evidence that pretraining design, not scale, drives performance}: w2v-BERT 2.0 (580M) outperforms larger multilingual models and comparable-scale baselines, indicating that pretraining design, not parameter count, drives low-resource dialectal performance.
\item \textbf{A characterisation of the residual error profile} (dominated by dependent vowel signs and conjunct marking), corroborated by independent analysis, indicating that the bottleneck is representational rather than procedural. We additionally include an exploratory, single-seed layer-wise probing analysis of where fine-tuning adapts the encoder.
\end{itemize}

\section{Related Work}
\label{sec:related}

\paragraph{Self-supervised speech models}
Self-supervised learning (SSL) has substantially advanced low-resource speech recognition by leveraging large-scale unlabeled audio during pretraining. wav2vec 2.0~\citep{baevski2020wav2vec2} showed that contrastive predictive learning over raw speech achieves strong ASR with minimal labeled data, and subsequent multilingual models extended this to strong cross-lingual transfer. XLS-R~\citep{babu2022xlsr} scaled wav2vec-style pretraining to hundreds of languages, while Massively Multilingual Speech (MMS)~\citep{pratap2024scaling} reached over 1{,}000 languages via multilingual pretraining and language-specific adapters. Our work builds on w2v-BERT 2.0~\citep{w2v}, which jointly optimizes contrastive learning and masked prediction in an end-to-end fashion, avoiding the iterative clustering of HuBERT~\citep{hsu2021hubert} and the separate quantization modules of vq-wav2vec~\citep{baevski2019vq}.

\paragraph{Low-resource and dialectal ASR}
Low-resource ASR remains challenging due to limited labeled data, acoustic variability, and the absence of standardized orthographies and large transcribed corpora~\citep{besacier2014automatic}. Within Indic speech technology, recent work has increasingly adopted multilingual and transfer-learning approaches~\citep{javed2022towards, pratap2020massively} and built large inclusive corpora~\citep{javed2024indicvoices}, yet dialectal varieties remain comparatively underexplored. The recent ``Dialect Matters'' study~\citep{dhasmana-etal-2026-dialect} examined cross-lingual transfer across Devanagari-script Indic varieties and, as a case study, provided the first Garhwali ASR benchmark, reporting 49.3\% WER with w2v-BERT 2.0 on an 87/6/7 split using single training runs. Their error analysis independently identifies vowel-length (matra), aspiration, and halant/word-boundary confusions as the dominant Garhwali error types, corroborating the residual-error pattern we analyse in Section~\ref{sec:error_analysis}. We build directly on this case study, but on the official VAANI splits and under a multi-seed protocol, and we investigate training objectives, augmentation, and internal representation dynamics, and test a single high-resource transfer source (Hindi) rather than the broad cross-lingual sweep across many languages that they perform. Table~\ref{tab:vs_prior} summarises how our benchmark differs along the axes most relevant to reproducibility; we compare directly against their reported numbers in Section~\ref{sec:reproducibility}, where the divergence between the two studies' baselines itself illustrates the single-run fragility we document.

\begin{table*}[t]
\centering
\setlength{\tabcolsep}{4pt}
\caption{How our Garhwali benchmark differs from the only prior Garhwali ASR results~\citep{dhasmana-etal-2026-dialect}, along the axes that govern reproducibility. The divergence between the two studies' baselines (Section~\ref{sec:reproducibility}) is itself a symptom of the single-run fragility these choices address.}
\label{tab:vs_prior}
\begin{tabular}{lcc}
\toprule
\textbf{Axis} & \textbf{\citet{dhasmana-etal-2026-dialect}} & \textbf{Ours} \\
\midrule
Split & train-internal 87/6/7 & official VAANI \\
Runs per system & single & 5 seeds \\
Significance test & no & yes (Holm) \\
Variance reported & no & s.d.\ + bootstrap \\
Per-seed release & no & yes \\
Early-stop metric & CER & WER (+CER check) \\
\bottomrule
\end{tabular}
\end{table*}

\paragraph{Focal loss and CTC-based ASR}
Connectionist Temporal Classification (CTC)~\citep{graves2006ctc} is widely used for end-to-end ASR as it learns speech-transcription alignments without frame-level supervision. However, standard CTC weights all training examples equally, over-emphasizing easy examples once they are well learned. Focal loss~\citep{lin2017focal}, proposed for dense object detection to down-weight easy examples and emphasize hard ones, has seen limited use in CTC-based ASR; the closest work applies focal weighting at the task level for multilingual meta-learning~\citep{chen2023task}, not to per-utterance CTC. Motivated by the difficulty imbalance in low-resource dialectal speech, we adapt focal modulation to sequence-level CTC; as our multi-seed evaluation shows (Section~\ref{sec:objective_comparison}), however, this objective-level intervention does not reliably improve over standard CTC, a result we frame as evidence that objective engineering is a less dependable lever than data augmentation in this regime.

\section{Experimental Setup}
\label{sec:methods}

\subsection{Data and Splits}
We use the Garhwali subset of the VAANI corpus~\citep{vaanicapturing2026} (5{,}894 utterances total)\footnote{\href{https://huggingface.co/datasets/ARTPARK-IISc/Vaani-transcription-part}{\texttt{ARTPARK-IISc/Vaani-transcription-part}}}, which ships with official train/validation/test splits of 4{,}778 / 666 / 450 utterances; the training split comprises 8.8\,h of transcribed field speech. Unlike prior work, which partitioned the training portion internally, we use these official splits throughout, so that all systems are compared on the same held-out test set and results are directly reproducible. Transcripts are normalised with a fixed pipeline (markup and zero-width-joiner removal, Latin-script stripping, whitespace collapsing); the resulting character set yields a 66-token Devanagari vocabulary (63 characters plus word-delimiter, unknown, and padding tokens); the tokenizer adds two sentence-boundary special tokens that never occur in any target, so the output layer has 68 logits. The same vocabulary and normalisation are used for every system in this paper.

\subsection{Multi-Seed Protocol}\label{sec:protocol}
All primary systems are trained over five random seeds (42, 123, 777, 2025, 1234). We report the mean and standard deviation of corpus-level WER and CER across seeds, and assess pairwise differences with paired Wilcoxon signed-rank tests on per-seed corpus WER, applying Holm--Bonferroni correction across the family of comparisons. We additionally report a secondary per-utterance bootstrap cross-check (Appendix~\ref{app:bootstrap}), noting that it understates uncertainty relative to the seed-level tests. One property of this design should be stated explicitly: with five paired seeds, the smallest attainable two-sided exact Wilcoxon $p$ is $0.0625$, and after a three-comparison Holm correction the smallest attainable adjusted value is $0.1875$. The seed-level tests therefore cannot reach $p<.05$ at this seed budget, even for a system that wins on all five seeds. We accordingly report them as a conservative descriptive summary, and we rest our conclusions on the paired per-seed differences, the effect sizes, and the power analysis of Section~\ref{sec:reproducibility}, which uses a paired $t$ test and has no such floor. This protocol is a deliberate departure from single-run reporting and follows recent calls to quantify seed- and data-induced variance rather than report a single run \citep{bouthillier2021accounting, bui2025assessing}: at this corpus size the per-utterance WER variance is large, and as we show in Section~\ref{sec:objective_comparison}, single-seed differences between objectives do not survive replication.

\subsection{Primary Model}
Our primary encoder is w2v-BERT 2.0 (\texttt{facebook/w2v-bert-2.0}),\footnote{\url{https://huggingface.co/facebook/w2v-bert-2.0}} a 580M-parameter multilingual speech model with 24 Transformer layers, based on the SeamlessM4T speech encoder. We fine-tune with CTC, freezing only the feature-projection layer and keeping the 24 encoder layers and the output head trainable. Inputs are 80-dimensional log-Mel features from the SeamlessM4T extractor at 16\,kHz. We use a dual learning rate: encoder $3\times10^{-5}$, head $10^{-3}$ with linear warmup over the first 10\% of steps then linear decay, AdamW, effective batch size 32, BF16 precision, up to 20 epochs with early stopping (patience 5) on validation WER. Final evaluation uses per-utterance greedy decoding.

\begin{figure*}[t]
    \centering
    \includegraphics[width=0.75\textwidth]{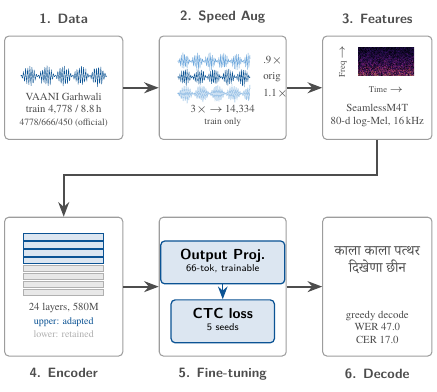}
\caption{The Garhwali ASR pipeline with w2v-BERT 2.0.
\textbf{(1)}~VAANI Garhwali (official 4{,}778/666/450 split; 8.8\,h train). \textbf{(2)}~$3\times$ speed augmentation ($0.9/1.0/1.1\times$) triples the training set to 14{,}334 (train only). \textbf{(3)}~80-dim log-Mel features (SeamlessM4T). \textbf{(4)}~24-layer w2v-BERT 2.0 encoder (580M), all layers fine-tuned; shading marks the depth-dependent adaptation found by probing (Appendix~\ref{app:probing}). \textbf{(5)}~A trainable 66-token CTC head; feature projection frozen; five seeds. \textbf{(6)}~Greedy decoding (standard CTC: 47.0/17.0 WER/CER).}
\label{fig:architecture}
\end{figure*}

\subsection{Training Objectives}
\label{sec:objectives}

We compare three CTC training objectives under an identical optimisation protocol, isolating the effect of the objective alone.

\paragraph{Standard CTC.}
The mean-reduced Connectionist Temporal Classification loss \citep{graves2006ctc}, marginalising over all monotonic alignments via the forward--backward algorithm. This is our floor.

\paragraph{Focal CTC.}
We adapt focal modulation \citep{lin2017focal} to the utterance level. Because CTC marginalises over all alignments, the raw per-utterance loss $\ell$ produces sequence-level probabilities so small that $\exp(-\ell)$ underflows to zero (empirically, $\exp(-340)=0$), which would make any confidence-based modulator a no-op. We therefore replace it with a length-normalised, temperature-scaled confidence $p$ (Appendix~\ref{app:objectives_math}, Eq.~\eqref{eq:focal_p}) and apply focal modulation to that:
\begin{equation}
  \mathcal{L}_{\text{focal}}
   \;=\; \operatorname{mean}\big[\,\alpha\,(1-p)^{\gamma}\,\ell\,\big],
  \label{eq:focal_loss}
\end{equation}
with $\alpha = 1.0$, $\gamma = 0.5$, and temperature $\tau = 10$. Setting $\gamma = 0$ recovers standard CTC exactly. The reduced $\gamma$ (versus the $\gamma = 2$ typical in image classification) and the temperature reflect the compressed dynamic range of length-normalised CTC confidences; both were fixed prior to the multi-seed comparison.

\paragraph{Matra-weighted CTC.}
To target the dominant residual error category (dependent vowel signs; Section~\ref{sec:error_analysis}) we add a phonologically-motivated \emph{auxiliary} term rather than reweighting the CTC loss itself. Since CTC does not factorise into per-character contributions, we keep the CTC objective intact and add a class-weighted auxiliary signal $\mathcal{L} = \mathcal{L}_{\text{base}} + \lambda\,\mathcal{L}_{\text{aux}}$ (Appendix~\ref{app:objectives_math}, Eqs.~\eqref{eq:matra_total}--\eqref{eq:matra_aux}), where $\mathcal{L}_{\text{base}}$ is the focal CTC loss of Eq.~\eqref{eq:focal_loss} and $\mathcal{L}_{\text{aux}}$ is a class-weighted negative log-likelihood on the per-frame posteriors. Each frame is assigned to the token the model emits under a cheap greedy best-path alignment; frames emitting phonologically salient tokens (matras, and optionally aspirated/retroflex consonants) are up-weighted by a class weight $w_c$, normalised over non-blank frames. We restrict the term to salient frames ($w_{c(f)} > 1$) so it does not compete with CTC on ordinary characters, and use $\lambda = 0.3$ with matra/aspirated/retroflex weights of $3.0/2.0/2.5$ (a character in multiple classes takes the maximum). We stress that $\mathcal{L}_{\text{aux}}$ is its own objective, not a decomposition of the CTC loss; this auxiliary alignment is greedy and approximate, adequate for an emphasis term and far cheaper than full forward--backward posterior alignment.

\subsection{Speed Augmentation}
\label{sec:speedaug}
We augment the training data with speed perturbation using \texttt{librosa} time-stretch at rates $0.9\times$ and $1.1\times$; combined with the unmodified ($1.0\times$) audio, this materialises three copies of each training utterance, tripling the set from 4{,}778 to 14{,}334 while leaving the transcripts unchanged. This increases acoustic diversity and robustness to speaking-rate variation~\citep{ko2015audio}. Augmentation is applied to the training split only; validation and test sets are never augmented.

\section{Results}
\label{sec:results}

\subsection{Multi-Seed Objective Comparison}
\label{sec:objective_comparison}

We evaluate three training objectives: standard CTC, Focal CTC, and a phonologically-motivated matra-weighted CTC each fine-tuned on the $3\times$ speed-augmented official training split across five random seeds (42, 123, 777, 2025, 1234). Reporting over multiple seeds, rather than a single run, is essential here: the per-utterance WER variance of short conversational Garhwali speech is high, and we find that single-seed differences between objectives do not survive replication.

\begin{table*}[t]
\centering
\setlength{\tabcolsep}{4pt}
\caption{Objective comparison on the official Garhwali test set, mean $\pm$ s.d.\ over five seeds. ``No Aug''/``+Aug'': WER without/with $3\times$ speed augmentation, $\Delta$ their difference; CER and matraER are for the augmented setting. matraER is the vowel-diacritic error rate the matra-weighted objective targets.}
\label{tab:objective_multiseed}
\begin{tabular}{lccccc}
\toprule
& \multicolumn{3}{c}{\textbf{WER}} & & \\
\cmidrule(lr){2-4}
\textbf{Objective} & \textbf{No Aug} & \textbf{+Aug} & \textbf{$\Delta$} & \textbf{CER} & \textbf{matraER} \\
\midrule
Standard CTC          & 48.10 \small$\pm$0.74 & \textbf{47.02 \small$\pm$0.61} & $-1.08$ & \textbf{16.99 \small$\pm$0.21} & 22.26 \\
Focal CTC             & 49.18 \small$\pm$1.17 & 47.83 \small$\pm$0.68 & $-1.35$ & 17.35 \small$\pm$0.30 & 22.78 \\
Matra-weighted (ours) & 49.00 \small$\pm$0.22 & 47.42 \small$\pm$0.78 & $-1.58$ & 17.22 \small$\pm$0.38 & 22.31 \\
\bottomrule
\end{tabular}
\end{table*}

Table~\ref{tab:objective_multiseed} shows that no objective improves on standard CTC, which attains the lowest mean WER (47.02), with Focal CTC (47.83) and the matra-weighted objective (47.42) within one standard deviation. Holm-corrected paired Wilcoxon tests on per-seed corpus WER find no significant pair (standard vs.\ focal, $p=0.19$; standard vs.\ matra, $p=0.38$; focal vs.\ matra, $p=0.38$); the first value is the five-seed floor of Section~\ref{sec:protocol}, reached because standard CTC beats focal on all five seeds (mean $+0.81$, $d_z=0.98$), while the standard-vs.-matra gap is small and sign-unstable. Any objective-level effect thus points toward the baseline, yet a single seed could easily suggest otherwise: the best Focal seed (46.77) and best standard seed (46.32) are within noise of one another. We therefore report multi-seed means with paired per-seed differences as the reliable characterisation.

\subsection{The Matra Objective Gains No Ground on Matra Errors}
\label{sec:matra_null}

The matra-weighted objective was designed specifically to address the dominant residual error category (vowel-diacritic confusions; see Section~\ref{sec:error_analysis}) by adding a matra-weighted auxiliary loss term (Section~\ref{sec:objectives}). It does not achieve this aim. Against standard CTC, the matra error rate is essentially unchanged (22.26\,$\pm$\,0.20\% vs.\ 22.31\,$\pm$\,0.77\%), a 0.05-point gap far smaller than the seed-to-seed spread. Against its actual base (the objective is built on Focal CTC, Eq.~\eqref{eq:matra_total}), it recovers most of the degradation that focal weighting introduces (22.78 to 22.31, lower on four of five seeds), but this reduction is within the seed spread and only returns the matra error rate to the standard-CTC level. Loss-level reweighting thus buys no ground on the targeted category, which likely reflects limits of the acoustic representation or the available data.

\subsection{Speed Augmentation Is the Most Consistent Lever}
\label{sec:augmentation_effect}

To isolate the contribution of speed augmentation, we retrain all three objectives without augmentation, again over five seeds; the comparison is the ``No Aug'' versus ``+Aug'' columns of Table~\ref{tab:objective_multiseed}.

The ``No Aug''/``+Aug'' columns of Table~\ref{tab:objective_multiseed} show speed augmentation lowers mean WER by 1.1--1.5 points under every objective. Held to the same standard as Section~\ref{sec:objective_comparison}, this is not statistically certified (exact paired Wilcoxon $p=0.125$, $0.125$, and $0.0625$, the last being the five-seed minimum), and 13 of 15 paired seeds improve, with two slight reversals (standard seed~777, focal seed~1234). It is nonetheless the most dependable lever we test: its mean effect is direction-consistent under all three objectives, whereas switching objectives moves WER by less than the seed-to-seed standard deviation.

\subsection{Cross-Lingual Transfer Gives No Reliable Gain}
\label{sec:transfer}

A natural alternative to direct fine-tuning is to transfer through a high-resource, phylogenetically related language. Hindi is the obvious candidate: it is the dominant Devanagari-script language, shares Garhwali's script, and is well represented in pretraining and fine-tuning resources. We therefore test a two-stage transfer: fine-tune w2v-BERT 2.0 on Hindi (FLEURS) and then on the official Garhwali training split, over the same five seeds and protocol as our primary systems.

Table~\ref{tab:transfer} shows that transferring through Hindi does not reliably improve over direct standard-CTC fine-tuning. The five-seed transfer mean is 47.22\,$\pm$\,0.91 WER, marginally \emph{worse} than the 47.02 of direct fine-tuning and not significantly different (paired Wilcoxon on per-seed WER, $p=0.81$; paired $t$, $p=0.64$). The per-seed differences change sign transfer helps on seeds~42 and~2025 but hurts on 123,~777, and~1234 so the Hindi stage neither helps nor hurts in a consistent direction.

This null is consistent with \citet{dhasmana-etal-2026-dialect}, who find that for dialectal Indic varieties, phylogenetic or script proximity to a high-resource language does not guarantee transfer benefit, fine-tuning on dialectal data directly is often as effective as routing through a related standardized language. Our result is the single-target, multi-seed analogue of that finding: for Garhwali specifically, a Hindi transfer stage is not a dependable lever, reinforcing our central theme that the reliable gains come from pretraining choice and augmentation rather than from additional training-pipeline machinery.

\begin{table*}[t]
\centering
\setlength{\tabcolsep}{5pt}
\caption{Hindi$\rightarrow$Garhwali two-stage transfer vs.\ direct standard-CTC fine-tuning, WER (\%) over five seeds; the lower value per row is in bold. Transfer is within noise of direct fine-tuning (paired Wilcoxon $p=0.81$) and the better system flips across seeds direct wins three, transfer two.}
\label{tab:transfer}
\begin{tabular}{lcc}
\toprule
\textbf{Seed} & \textbf{Direct (standard CTC)} & \textbf{Hindi transfer} \\
\midrule
42   & 46.73 & \textbf{46.33} \\
123  & \textbf{46.32} & 47.80 \\
777  & \textbf{47.93} & 48.46 \\
2025 & 47.25 & \textbf{46.42} \\
1234 & \textbf{46.88} & 47.10 \\
\midrule
Mean & \textbf{47.02 \small$\pm$0.61} & 47.22 \small$\pm$0.91 \\
\bottomrule
\end{tabular}
\end{table*}

\subsection{Comparison with Prior Garhwali ASR}
\label{sec:reproducibility}

\paragraph{Comparison with \citet{dhasmana-etal-2026-dialect}.}
The same setup-sensitivity appears when we compare against the only prior Garhwali ASR numbers. \citet{dhasmana-etal-2026-dialect} fine-tune the same encoder (w2v-BERT 2.0 with CTC) but on a train-internal 87/6/7 split, with single runs and CER-based early stopping; their best system reaches 49.3 WER. Our five-seed mean on the official split is 47.0\,$\pm$\,0.6. More striking is that the \emph{other} encoders disagree sharply across the two studies: they report XLS-R at 65.0 and HuBERT at 51.5 WER, whereas we obtain 50.2 and 60.9 respectively, the two systems swap relative order between the studies. These discrepancies are not contradictions but a direct illustration of our central point: at this corpus size, single-run dialectal ASR numbers are highly sensitive to the choice of split, early-stopping metric, and seed, so encoder rankings drawn from single runs need not replicate. Reporting on the official splits over multiple seeds is precisely what makes such comparisons stable.

We also verify the gap is not an artefact of our early-stopping criterion: re-running standard CTC with CER-based stopping (as in \citealp{dhasmana-etal-2026-dialect}) gives 47.56$\pm$0.74 WER, within noise of our WER-stopped 47.02$\pm$0.61 (Appendix~\ref{app:repro}), so the 47-vs-49 gap is attributable to the split and single-run reporting, not the stopping metric.

\paragraph{Is the null result a power problem?}
The seed-level Wilcoxon tests are floored at five seeds (Section~\ref{sec:protocol}), and even an unfloored test may lack power to resolve a small real effect. We therefore report a post-hoc power analysis on the per-seed corpus WER using a two-sided paired $t$ test, which has no such floor (Table~\ref{tab:power}). The objective gaps are not negligible in standardised terms: the largest, standard vs.\ Focal CTC, has a paired effect size of $d_z = 0.98$ but with five seeds the achieved power to detect it is only $0.39$ at $\alpha = 0.05$, and approximately $11$ seeds would be required for $80\%$ power. The standard vs.\ matra-weighted gap ($d_z = 0.66$) would need roughly $20$ seeds. We therefore characterise the objective differences as \emph{not reliably established at a practical seed budget} rather than as demonstrably zero. Crucially, even the largest gap is under one WER point in absolute terms, and where it would become significant (standard vs.\ Focal) the direction favours \emph{standard} CTC that is, the additional seeds would confirm that the focal objective slightly \emph{hurts}, not that it helps. This reinforces, rather than weakens, our central claim: objective-level interventions are not a dependable lever in this setting, whereas speed augmentation improves the mean WER across all objectives (Section~\ref{sec:augmentation_effect}).

\begin{table*}[t]
\centering
\setlength{\tabcolsep}{4pt}
\caption{Post-hoc power on per-seed corpus WER (five seeds). $d_z$: paired effect size; power@5: achieved power at $n{=}5$, $\alpha{=}0.05$ (two-sided paired $t$); $n_{80}$: seeds needed for $80\%$ power.}
\label{tab:power}
\begin{tabular}{lrrrr}
\toprule
\textbf{Pair} & \textbf{$\Delta$WER} & \textbf{$d_z$} & \textbf{power@5} & \textbf{$n_{80}$} \\
\midrule
standard vs.\ focal & $+0.81$ & $0.98$ & $0.39$ & $11$ \\
standard vs.\ matra & $+0.40$ & $0.66$ & $0.21$ & $20$ \\
focal vs.\ matra    & $-0.42$ & $0.91$ & $0.34$ & $12$ \\
\bottomrule
\end{tabular}
\end{table*}

\section{Analysis}
\label{sec:analysis}

\subsection{Error Analysis}
\label{sec:error_analysis}

To characterise \emph{which} errors dominate and whether the objective-level interventions shift the error profile, we compute per-category error rates over phonologically defined character classes, aggregated over the same five seeds as our main results. For each class we report the fraction of reference characters in that class that are misrecognised; Table~\ref{tab:class_errors} summarises the three objectives. Table~\ref{app:tab:examples} complements this quantitative view with representative transcriptions sampled across the full per-utterance error range (0--50\% CER), showing how recognition errors manifest in practice from exact matches and local orthographic/phonetic deviations to heavily degraded output.

\begin{table*}[t]
\centering
\setlength{\tabcolsep}{4pt}
\caption{Per-category character error rate (\%) on the official test set, mean $\pm$ s.d.\ over five seeds, by objective. Categories are phonological character classes; rates count substitutions and deletions only (insertions have no reference position), so ``Categorized'' is below the corpus CER of Table~\ref{tab:objective_multiseed}. $^\dagger$Nukta ($n{=}253$): high-variance, not interpreted.}
\label{tab:class_errors}
\begin{tabular}{lccc}
\toprule
\textbf{Category} & \textbf{Standard} & \textbf{Focal} & \textbf{Matra} \\
\midrule
Virama / halant      & 29.51 \small$\pm$2.59 & 32.40 \small$\pm$3.59 & 29.56 \small$\pm$1.95 \\
Nasal                & 25.56 \small$\pm$2.46 & 26.77 \small$\pm$1.27 & 26.17 \small$\pm$2.12 \\
Indep.\ vowel        & 23.03 \small$\pm$1.37 & 23.27 \small$\pm$1.65 & 22.65 \small$\pm$1.35 \\
Matra                & \textbf{22.26 \small$\pm$0.20} & 22.78 \small$\pm$0.55 & 22.31 \small$\pm$0.77 \\
Aspirated            & 15.45 \small$\pm$1.88 & 14.72 \small$\pm$1.05 & 15.16 \small$\pm$0.94 \\
Retroflex            & 10.71 \small$\pm$0.52 & 10.83 \small$\pm$0.43 & 10.47 \small$\pm$0.63 \\
Nukta$^\dagger$      & 12.89 \small$\pm$6.22 & 17.63 \small$\pm$8.49 & 18.74 \small$\pm$6.97 \\
\midrule
Categorized (sub+del) & 16.07 \small$\pm$0.30 & 16.69 \small$\pm$0.57 & 16.25 \small$\pm$0.44 \\
\bottomrule
\end{tabular}
\end{table*}

\paragraph{The error profile is dominated by dependent signs and conjuncts.}
Across all objectives the highest error rates fall on virama/halant ($\sim$30\%), nasal marking ($\sim$26\%), and the two vowel categories ($\sim$22--23\%), while retroflex and aspirated consonants are far more reliable ($\sim$10--15\%). This pattern matches the independent error analysis of \citet{dhasmana-etal-2026-dialect}, whose dominant errors are also word-boundary confusions, vowel-length (\dev{ु} vs.\ \dev{ू}), halant deletion, and aspiration (\dev{ब} vs.\ \dev{भ}). That two studies on different splits and protocols surface the same structure corroborates that these categories are intrinsic to the task, not artefacts of our setup.

\paragraph{Objective changes do not improve the error profile.}
Neither objective-level intervention produces a favourable shift. The matra-weighted objective leaves its target category unchanged (\S\ref{sec:matra_null}), and Focal CTC is, if anything, counterproductive: it has the highest error rate in nearly every category, including the dominant virama category (32.40\% vs.\ 29.51\% for standard) and the aggregate categorized rate (16.69\% vs.\ 16.07\%). The residual error structure is thus stable under loss reweighting, with the dominant errors in dependent vowel signs and conjunct marking phenomena more plausibly limited by acoustic and data evidence than by the objective.

\paragraph{The bottleneck is representational, not procedural.} Read together, our results point away from the training pipeline and toward the representation: the residual errors lie in fine phonological contrasts vowel length, aspiration, conjunct marking that loss reweighting cannot sharpen and a related standard language does not supply. Loss design and transfer can only redistribute information the model already has; for Garhwali, the limiting factor appears to be how much disambiguating evidence the acoustic representation and the available data contain.

\subsection{Cross-Model Error Categorisation}
\label{sec:llm_error}

A single-seed, four-model error categorisation of the standard-CTC predictions (Appendix~\ref{app:llm_error}), used here only as a directional cross-check on the seed-aggregated profile above, agrees on three points: orthographic/phonetic drift dominates the errors; apparent dialect normalisation is mostly phonetic drift on Hindi-looking spellings rather than true grammatical normalisation; and the system hallucinates text on every true-silence segment ($8/8$). The first two reinforce the representational-bottleneck reading; the third is a reliable failure our character-class profile does not surface, and points to voice-activity gating as a low-cost mitigation.

\paragraph{Scale does not predict performance.}
The 580M w2v-BERT 2.0 encoder outperforms the larger MMS-1B and comparable-scale models (XLS-R, HuBERT; Appendix~\ref{app:perseed}, Table~\ref{app:tab:baseline_perseed} and Figure~\ref{fig:scale}): the gap tracks pretraining design and coverage, not parameter count. Whisper Large-v3, a reference point only, is both worse and unstable under forced-Hindi decoding (two of five seeds failed; reported over the three valid seeds), reinforcing the point that for low-resource deployment the right pretrained encoder matters more than the largest one.

\paragraph{The cost of single-seed reporting.}
The fragility we document is not specific to one intervention: focal CTC interleaves with standard CTC across seeds and shows no reliable gain (Section~\ref{sec:objective_comparison}), and encoder rankings likewise differ across single-run studies (Section~\ref{sec:reproducibility}). At these corpus sizes, per-run variance is large enough to manufacture effects that do not survive replication which is why we argue multi-seed reporting with significance testing should be standard practice for this class of problem.

\subsection{What This Study Establishes}
\label{sec:takeaway}

This study contributes three results new to Garhwali ASR. First, a reproducible multi-seed benchmark on the official VAANI splits with per-seed outputs, variance estimates, and significance tests makes claims in this setting checkable for the first time. Second, against that benchmark the candidate interventions separate cleanly: pretraining choice improves WER by margins far exceeding seed noise, speed augmentation improves mean WER under every objective, and focal weighting, the matra objective, and Hindi transfer show no consistent paired advantage over standard CTC. Third, the residual difficulty concentrates on dependent vowel signs and conjunct marking, is stable across objectives, and is corroborated by independent analysis. The unifying lesson is methodological: the multi-seed discipline that overturns these specific claims is the minimum any future dialectal-ASR result needs to be trusted.

\section{Conclusion}
\label{sec:conclusion}

Our study set out to determine which reported and plausible gains for low-resource Garhwali ASR survive rigorous evaluation. The answer is sobering: under multi-seed testing on the official splits, the objective-level and transfer interventions we examined dissolve into seed noise, and even an objective built specifically to attack the dominant error category fails to move it. What remains robust is mundane, a strong pretrained encoder and simple speed augmentation. We read this as evidence that, for dialectal ASR at this corpus scale, the limiting factor is the acoustic and data evidence available to the model, not the training objective: loss design can only redistribute information the representation already contains. The methodological corollary is that single-seed results in this regime should not be trusted as evidence of an effect. For Garhwali specifically, the persistence of matra and conjunct errors under every intervention we tried suggests the next gains will come from richer acoustic data and better representations, not from the loss function.

\section*{Limitations}

Our conclusions are scoped to the settings we evaluate, and several limits qualify them. First, the study covers a single dialect (Garhwali) on a single corpus (the VAANI Garhwali subset, 8.8\,h of training audio); whether the same \emph{ordering} of levers pretraining and augmentation over objective and transfer engineering holds for other low-resource varieties is an empirical question we do not settle here. The methodological concern, however, is not Garhwali-specific: whenever seed-to-seed variability is comparable to the effect sizes being claimed, single-run comparisons can be misleading. The multi-seed protocol used here, including variance reporting and paired statistical assessment, can be applied directly to other low-resource benchmarks to test whether reported gains are robust. Second, our negative results are statistical, not absolute: with five seeds the objective and transfer comparisons are \emph{not reliably established at a practical seed budget} rather than demonstrably zero, and the post-hoc power analysis (Section~\ref{sec:reproducibility}) indicates the only gap likely to reach significance with more seeds (standard vs.\ Focal) favours standard CTC. Third, we test one transfer source (Hindi) and a specific pair of objective variants; other source languages, multi-source transfer, or alternative objective formulations could behave differently. We also do not evaluate parameter-efficient adaptation (e.g., adapters or LoRA) or self-training on unlabelled Garhwali audio. These are complementary strategies and natural candidates for evaluation under the same multi-seed protocol in future work. Fourth, the layer-wise probing and per-category error analyses are mechanistic and exploratory rather than confirmatory; the probing in particular is single-seed, and where analyses rely on automated alignment we treat conclusions as tentative and anchor claims to objective signals where possible. Finally, all CTC systems use greedy decoding without an external language model; Whisper, a reference point only, decodes by generation. This is a deliberate design choice rather than an oversight: holding the decoder fixed across systems isolates differences in the training setup encoder, objective, and augmentation from variation introduced by decoder-specific tuning. Our comparisons therefore concern differences in the acoustic-model training setup under a common decoding protocol. Whether beam search or LM-fused decoding reduces the residual errors observed here, or changes the relative ordering of systems, is a complementary question we leave open.

\section*{Ethics Statement}

The VAANI corpus is released by its authors for research use; we use only the publicly distributed Garhwali subset under its accompanying terms and add no new recordings, so the consent and collection protocols are those of the original corpus. We report aggregate error metrics only and present no speaker-identifying content. We note two risks specific to dialectal ASR. First, systems trained on a single standardized orthography can implicitly privilege one written norm for a language with substantial regional variation; our normalisation pipeline is a modelling convenience, not a prescriptive standard. Second, at the error rates reported here (best WER $\sim$47\%), the system is well below deployment quality, and we caution against using it in settings such as automated transcription of official, legal, or medical speech where recognition errors could disadvantage speakers of an already under-resourced language, a risk documented for Indian-language ASR in clinical settings \citep{kumar2025asr}. Our intended contribution is a reproducible benchmark to support further research, not a production system. We release code, splits, and per-seed outputs to support reproducibility and equitable progress on Garhwali and related varieties.

\paragraph{Use of AI assistants.}
Except for the explicitly described LLM-assisted error categorisation in Appendix~\ref{app:llm_error}, the authors used ChatGPT solely for editorial assistance, including improvements to clarity, grammar, conciseness, and stylistic consistency. Generative AI was not used to originate the research questions, experimental design, experimental results, or scientific conclusions. In Appendix~\ref{app:llm_error}, LLMs were used as an analysis instrument for the cross-model error categorisation, following the procedure described there. All AI-assisted content and analyses were reviewed and verified by the authors, who take full responsibility for the manuscript and its scientific claims.

\section*{Acknowledgements}
The authors are grateful for access to the Tier 2 High-Performance Computing resources provided by the Northern Ireland High-Performance Computing (NIHPC) facility, funded by the UK Engineering and Physical Sciences Research Council (EPSRC), Grant No. EP/T022175/1.

\bibliography{custom}

\begin{thebibliography}{22}
\providecommand{\natexlab}[1]{#1}

\bibitem[{Babu et~al.(2022)Babu, Wang, Tjandra, Lakhotia, Xu, Goyal, Singh,
  {von Platen}, Saraf, Pino, Baevski, Conneau, and Auli}]{babu2022xlsr}
Arun Babu, Changhan Wang, Andros Tjandra, Kushal Lakhotia, Qiantong Xu, Naman
  Goyal, Kritika Singh, Patrick {von Platen}, Yatharth Saraf, Juan Pino, Alexei
  Baevski, Alexis Conneau, and Michael Auli. 2022.
\newblock \href {https://doi.org/10.21437/Interspeech.2022-143} {{XLS-R:
  Self-supervised Cross-lingual Speech Representation Learning at Scale}}.
\newblock In \emph{{Interspeech 2022}}, pages 2278--2282.

\bibitem[{Baevski et~al.(2019)Baevski, Schneider, and Auli}]{baevski2019vq}
Alexei Baevski, Steffen Schneider, and Michael Auli. 2019.
\newblock vq-wav2vec: Self-supervised learning of discrete speech
  representations.
\newblock \emph{arXiv preprint arXiv:1910.05453}.

\bibitem[{Baevski et~al.(2020)Baevski, Zhou, Mohamed, and
  Auli}]{baevski2020wav2vec2}
Alexei Baevski, Yuhao Zhou, Abdelrahman Mohamed, and Michael Auli. 2020.
\newblock wav2vec 2.0: A framework for self-supervised learning of speech
  representations.
\newblock \emph{Advances in neural information processing systems},
  33:12449--12460.

\bibitem[{Besacier et~al.(2014)Besacier, Barnard, Karpov, and
  Schultz}]{besacier2014automatic}
Laurent Besacier, Etienne Barnard, Alexey Karpov, and Tanja Schultz. 2014.
\newblock Automatic speech recognition for under-resourced languages: A survey.
\newblock \emph{Speech communication}, 56:85--100.

\bibitem[{Bouthillier et~al.(2021)Bouthillier, Delaunay, Bronzi, Trofimov,
  Nichyporuk, Szeto, Mohammadi~Sepahvand, Raff, Madan, Voleti, Ebrahimi~Kahou,
  Michalski, Arbel, Pal, Varoquaux, and Vincent}]{bouthillier2021accounting}
Xavier Bouthillier, Pierre Delaunay, Mirko Bronzi, Assya Trofimov, Brennan
  Nichyporuk, Justin Szeto, Nazanin Mohammadi~Sepahvand, Edward Raff, Kanika
  Madan, Vikram Voleti, Samira Ebrahimi~Kahou, Vincent Michalski, Tal Arbel,
  Chris Pal, Gael Varoquaux, and Pascal Vincent. 2021.
\newblock \href
  {https://proceedings.mlsys.org/paper_files/paper/2021/file/0184b0cd3cfb185989f858a1d9f5c1eb-Paper.pdf}
  {Accounting for variance in machine learning benchmarks}.
\newblock In \emph{Proceedings of Machine Learning and Systems}, volume~3,
  pages 747--769.

\bibitem[{Bui et~al.(2025)Bui, Savova, and Wang}]{bui2025assessing}
Nghia~Tuan Bui, Guergana~K Savova, and Lijing Wang. 2025.
\newblock Assessing the macro and micro effects of random seeds on fine-tuning
  large language models.
\newblock In \emph{Proceedings of the 14th International Joint Conference on
  Natural Language Processing and the 4th Conference of the Asia-Pacific
  Chapter of the Association for Computational Linguistics}, pages 41--46.

\bibitem[{Chen et~al.(2023)Chen, Zhang, Zhang, Qu, and Yang}]{chen2023task}
Yaqi Chen, Wenlin Zhang, Hao Zhang, Dan Qu, and Xu-Kui Yang. 2023.
\newblock Task-based meta focal loss for multilingual low-resource speech
  recognition.
\newblock \emph{ACM Transactions on Asian and Low-Resource Language Information
  Processing}, 22(11):1--17.

\bibitem[{Chung et~al.(2021)Chung, Zhang, Han, Chiu, Qin, Pang, and Wu}]{w2v}
Yu-An Chung, Yu~Zhang, Wei Han, Chung-Cheng Chiu, James Qin, Ruoming Pang, and
  Yonghui Wu. 2021.
\newblock \href {https://doi.org/10.1109/ASRU51503.2021.9688253} {w2v-bert:
  Combining contrastive learning and masked language modeling for
  self-supervised speech pre-training}.
\newblock In \emph{2021 IEEE Automatic Speech Recognition and Understanding
  Workshop (ASRU)}, pages 244--250.

\bibitem[{Dhasmana et~al.(2026)Dhasmana, Srivastava, and
  Chiang}]{dhasmana-etal-2026-dialect}
Akriti Dhasmana, Aarohi Srivastava, and David Chiang. 2026.
\newblock \href {https://doi.org/10.18653/v1/2026.vardial-1.12} {Dialect
  matters: Cross-lingual {ASR} transfer for low-resource {I}ndic language
  varieties}.
\newblock In \emph{Proceedings of the 13th Workshop on {NLP} for Similar
  Languages, Varieties and Dialects}, pages 145--156, Rabat, Morocco.
  Association for Computational Linguistics.

\bibitem[{Graves et~al.(2006)Graves, Fernandez, Gomez, and
  Schmidhuber}]{graves2006ctc}
Alex Graves, Santiago Fernandez, Faustino Gomez, and J{\"u}rgen Schmidhuber.
  2006.
\newblock Connectionist temporal classification: Labelling unsegmented sequence
  data with recurrent neural networks.
\newblock In \emph{Proc. ICML}, pages 369--376.

\bibitem[{Hsu et~al.(2021)Hsu, Bolte, Tsai, Lakhotia, Salakhutdinov, and
  Mohamed}]{hsu2021hubert}
Wei-Ning Hsu, Benjamin Bolte, Yao-Hung~Hubert Tsai, Kushal Lakhotia, Ruslan
  Salakhutdinov, and Abdelrahman Mohamed. 2021.
\newblock \href {https://doi.org/10.1109/TASLP.2021.3122291} {Hubert:
  Self-supervised speech representation learning by masked prediction of hidden
  units}.
\newblock \emph{IEEE/ACM Trans. Audio, Speech and Lang. Proc.}, 29:3451–3460.

\bibitem[{Javed et~al.(2022)Javed, Doddapaneni, Raman, Bhogale, Ramesh,
  Kunchukuttan, Kumar, and Khapra}]{javed2022towards}
Tahir Javed, Sumanth Doddapaneni, Abhigyan Raman, Kaushal~Santosh Bhogale,
  Gowtham Ramesh, Anoop Kunchukuttan, Pratyush Kumar, and Mitesh~M Khapra.
  2022.
\newblock Towards building asr systems for the next billion users.
\newblock In \emph{Proceedings of the aaai conference on artificial
  intelligence}, volume~36, pages 10813--10821.

\bibitem[{Javed et~al.(2024)Javed, Nawale, George, Joshi, Bhogale, Mehendale,
  Sethi, Ananthanarayanan, Faquih, Palit, Ravishankar, Sukumaran, Panchagnula,
  Murali, Gandhi, R, M, Vaijayanthi, Karunganni, Kumar, and
  Khapra}]{javed2024indicvoices}
Tahir Javed, Janki Nawale, Eldho George, Sakshi Joshi, Kaushal Bhogale, Deovrat
  Mehendale, Ishvinder Sethi, Aparna Ananthanarayanan, Hafsah Faquih, Pratiti
  Palit, Sneha Ravishankar, Saranya Sukumaran, Tripura Panchagnula, Sunjay
  Murali, Kunal Gandhi, Ambujavalli R, Manickam M, C~Vaijayanthi, Krishnan
  Karunganni, and 2 others. 2024.
\newblock \href {https://doi.org/10.18653/v1/2024.findings-acl.639}
  {{I}ndic{V}oices: Towards building an inclusive multilingual speech dataset
  for {I}ndian languages}.
\newblock In \emph{Findings of the Association for Computational Linguistics:
  ACL 2024}, pages 10740--10782, Bangkok, Thailand. Association for
  Computational Linguistics.

\bibitem[{K et~al.(2025)K, James, Gopinath, and K}]{thennal2025advocating}
Thennal~D K, Jesin James, Deepa~Padmini Gopinath, and Muhammed~Ashraf K. 2025.
\newblock \href {https://doi.org/10.18653/v1/2025.findings-naacl.277}
  {Advocating character error rate for multilingual {ASR} evaluation}.
\newblock In \emph{Findings of the Association for Computational Linguistics:
  NAACL 2025}, pages 4941--4950, Albuquerque, New Mexico. Association for
  Computational Linguistics.

\bibitem[{Ko et~al.(2015)Ko, Peddinti, Povey, Seltzer, and
  Khudanpur}]{ko2015audio}
Tom Ko, Vijayaditya Peddinti, Daniel Povey, Michael~L. Seltzer, and Sanjeev
  Khudanpur. 2015.
\newblock Audio augmentation for speech recognition.
\newblock In \emph{Proc. Interspeech}, pages 3586--3589.

\bibitem[{Kumar et~al.(2025)Kumar, Shivaprakash, Manoharan, Kurariya,
  Mukherjee, Shukla, Mukherjee, Chand, and Murthy}]{kumar2025asr}
Subham Kumar, Prakrithi Shivaprakash, Abhishek Manoharan, Astut Kurariya,
  Diptadhi Mukherjee, Lekhansh Shukla, Animesh Mukherjee, Prabhat Chand, and
  Pratima Murthy. 2025.
\newblock Asr under the stethoscope: Evaluating biases in clinical speech
  recognition across indian languages.
\newblock \emph{arXiv preprint arXiv:2512.10967}.

\bibitem[{Lin et~al.(2017)Lin, Goyal, Girshick, He, and Dollar}]{lin2017focal}
Tsung-Yi Lin, Priya Goyal, Ross Girshick, Kaiming He, and Piotr Dollar. 2017.
\newblock Focal loss for dense object detection.
\newblock In \emph{IEEE International Conference on Computer Vision (ICCV)},
  pages 2980--2988.

\bibitem[{Picard(2021)}]{picard2021torch}
David Picard. 2021.
\newblock Torch. manual\_seed (3407) is all you need: On the influence of
  random seeds in deep learning architectures for computer vision.
\newblock \emph{arXiv preprint arXiv:2109.08203}.

\bibitem[{Pratap et~al.(2020)Pratap, Sriram, Tomasello, Hannun, Liptchinsky,
  Synnaeve, and Collobert}]{pratap2020massively}
Vineel Pratap, Anuroop Sriram, Paden Tomasello, Awni Hannun, Vitaliy
  Liptchinsky, Gabriel Synnaeve, and Ronan Collobert. 2020.
\newblock \href {https://doi.org/10.21437/Interspeech.2020-2831} {{Massively
  Multilingual ASR: 50 Languages, 1 Model, 1 Billion Parameters}}.
\newblock In \emph{{Interspeech 2020}}, pages 4751--4755.

\bibitem[{Pratap et~al.(2024)Pratap, Tjandra, Shi, Tomasello, Babu, Kundu,
  Elkahky, Ni, Vyas, Fazel-Zarandi, Baevski, Adi, Zhang, Hsu, Conneau, and
  Auli}]{pratap2024scaling}
Vineel Pratap, Andros Tjandra, Bowen Shi, Paden Tomasello, Arun Babu, Sayani
  Kundu, Ali Elkahky, Zhaoheng Ni, Apoorv Vyas, Maryam Fazel-Zarandi, Alexei
  Baevski, Yossi Adi, Xiaohui Zhang, Wei-Ning Hsu, Alexis Conneau, and Michael
  Auli. 2024.
\newblock Scaling speech technology to 1,000+ languages.
\newblock \emph{Journal of Machine Learning Research}, 25(97):1--52.

\bibitem[{Pulikodan et~al.(2026)Pulikodan, Singh, Basu, Desai, J, Bhat,
  Dharmaraju, Gupta, Udupa, Kumar, Sharma, Sanka, Tewari, Dhand, Kamat, Singh,
  Vashishth, Talukdar, Acharya, and Ghosh}]{vaanicapturing2026}
Sujith Pulikodan, Abhayjeet Singh, Agneedh Basu, Nihar Desai, Pavan~Kumar J,
  Pranav~D Bhat, Raghu Dharmaraju, Ritika Gupta, Sathvik Udupa, Saurabh Kumar,
  Sumit Sharma, Visruth Sanka, Dinesh Tewari, Harsh Dhand, Amrita Kamat,
  Sukhwinder Singh, Shikhar Vashishth, Partha Talukdar, Raj Acharya, and
  Prasanta~Kumar Ghosh. 2026.
\newblock \href {https://arxiv.org/abs/2603.28714} {Vaani: Capturing the
  language landscape for an inclusive digital india}.
\newblock \emph{Preprint}, arXiv:2603.28714.

\bibitem[{Srivastav et~al.(2025)Srivastav, Zheng, Bezzam, Bihan, Moumen, and
  Gandhi}]{srivastav2025open}
Vaibhav Srivastav, Steven Zheng, Eric Bezzam, Eustache~Le Bihan, Adel Moumen,
  and Sanchit Gandhi. 2025.
\newblock Open asr leaderboard: Towards reproducible and transparent
  multilingual speech recognition evaluation.
\newblock \emph{arXiv preprint arXiv:2510.06961}.

\end{thebibliography}

\appendix

\section*{Appendix}

\noindent
This appendix provides the complete evidence behind the aggregate numbers in the main paper. Because our central claim is methodological that single-seed comparisons are unreliable at this corpus size, we report every per-seed value rather than means alone, so that the seed-to-seed spread is fully visible and the analysis is independently reproducible. Appendix~\ref{app:perseed} gives per-seed WER/CER for every system; Appendix~\ref{app:classerr} breaks the character errors down by phonological category for every seed and condition; and Appendix~\ref{app:repro} lists the shared training configuration and the full significance results.

\section{Per-Seed Results}
\label{app:perseed}

The tables below expand every WER/CER mean reported in the body into its five constituent seeds. They make two patterns directly inspectable. First, the seed-to-seed spread within a system is comparable to often larger than the gap \emph{between} objectives: in Table~\ref{app:tab:obj_perseed} the standard CTC seeds range from 46.32 to 47.93 WER, a 1.6-point spread that comfortably straddles the 0.4--0.8-point mean differences between objectives. This is the empirical basis for our argument that a single seed can easily misrank the three objectives. Second, the ordering of seeds is not stable across systems (the best seed for one objective is not the best for another), which is why a paired test across shared seeds, rather than a comparison of best runs, is the appropriate analysis. Throughout, cells show WER/CER (\%) on the official VAANI test set (450 utterances) and ``--'' marks a seed with no valid run (see table footnotes). Seeds are 42, 123, 777, 2025, 1234.

\begin{table*}[t]
\centering\small\setlength{\tabcolsep}{3pt}
\caption{Per-seed WER/CER (\%) for the three objectives, with and without $3\times$ speed augmentation (official test set, 450 utterances). Final column is WER mean$\pm$s.d.; the augmented block matches Table~\ref{tab:objective_multiseed}.}
\label{app:tab:obj_perseed}
\begin{tabular}{lcccccc}
\toprule
\textbf{System} & \textbf{s42} & \textbf{s123} & \textbf{s777} & \textbf{s2025} & \textbf{s1234} & \textbf{Mean$\pm$s.d.} \\
\midrule
\multicolumn{7}{l}{\emph{With $3\times$ speed augmentation}} \\
\midrule
Standard CTC & 46.73/16.71 & 46.32/17.00 & 47.93/17.25 & 47.25/16.87 & 46.88/17.12 & 47.02\,$\pm$\,0.61 \\
Focal CTC & 46.77/16.89 & 48.40/17.66 & 48.36/17.45 & 47.58/17.27 & 48.06/17.50 & 47.83\,$\pm$\,0.68 \\
Matra-weighted & 46.52/16.98 & 47.48/16.96 & 48.65/17.80 & 47.05/16.96 & 47.40/17.42 & 47.42\,$\pm$\,0.78 \\
\midrule
\multicolumn{7}{l}{\emph{Without augmentation}} \\
\midrule
Standard CTC & 47.50/17.30 & 49.28/18.16 & 47.61/17.53 & 48.36/17.74 & 47.76/17.40 & 48.10\,$\pm$\,0.74 \\
Focal CTC & 48.46/17.55 & 49.19/17.96 & 50.89/19.04 & 49.54/18.35 & 47.80/17.43 & 49.18\,$\pm$\,1.17 \\
Matra-weighted & 49.19/17.98 & 48.73/17.69 & 48.84/18.11 & 49.23/17.65 & 49.03/17.58 & 49.00\,$\pm$\,0.22 \\
\bottomrule
\end{tabular}
\end{table*}

\noindent
Table~\ref{app:tab:baseline_perseed} reports the comparison encoders.\footnote{HuggingFace checkpoints:
\href{https://huggingface.co/facebook/mms-1b-all}{\texttt{facebook/mms-1b-all}},
\href{https://huggingface.co/facebook/wav2vec2-xls-r-300m}{\texttt{facebook/wav2vec2-xls-r-300m}},
\href{https://huggingface.co/facebook/hubert-large-ll60k}{\texttt{facebook/hubert-large-ll60k}},
\href{https://huggingface.co/openai/whisper-large-v3}{\texttt{openai/whisper-large-v3}}.} The
within-system seed spreads here are small (s.d.\ $\le$0.5 WER for the self-supervised CTC baselines), so the large gaps between encoders: w2v-BERT 2.0 at 47.0, MMS-1B at 49.0, XLS-R~300M at 50.2, HuBERT at 60.9 are robust and not seed artefacts. Whisper Large-v3 is the exception in both level and stability: its valid seeds span 63--70 WER and two seeds failed outright (walltime and degenerate decode), reflecting the fragility of forced-Hindi decoding on noisy dialectal audio. Where individual seeds failed to converge (HuBERT seed~777) or failed outright (two Whisper seeds), we report the converged seeds and footnote the exclusions rather than dropping them silently.

\begin{table*}[t]
\centering\small\setlength{\tabcolsep}{3pt}
\caption{Per-seed WER/CER for baseline encoders. Means are over available seeds; incomplete seed sets are footnoted. The between-encoder gaps far exceed the within-encoder seed spread, except for Whisper, which is unstable in both level and variance.}
\label{app:tab:baseline_perseed}
\begin{tabular}{lcccccc}
\toprule
\textbf{System} & \textbf{s42} & \textbf{s123} & \textbf{s777} & \textbf{s2025} & \textbf{s1234} & \textbf{Mean$\pm$s.d.} \\
\midrule
MMS-1B & 49.29/17.89 & 48.83/17.80 & 48.18/17.54 & 49.19/18.04 & 49.39/17.83 & 48.98\,$\pm$\,0.49 \\
XLS-R 300M & 50.66/18.71 & 50.07/18.58 & 50.14/18.56 & 50.26/18.69 & 50.01/18.46 & 50.23\,$\pm$\,0.26 \\
HuBERT$^{a}$ & 61.50/22.98 & 60.33/22.71 & 67.12/25.56 & 60.73/22.83 & 61.05/22.71 & 60.90\,$\pm$\,0.49 \\
Whisper Large-v3$^{b}$ & 65.90/39.69 & 70.20/38.84 & 62.96/29.97 & -- & -- & 66.35\,$\pm$\,3.64 \\
\bottomrule
\end{tabular}
\\[2pt]\footnotesize
$^{a}$HuBERT seed~777 (shown in grey) failed to converge (val WER 0.73 vs.\ $\sim$0.66 for the other seeds; test WER 67.1) and is excluded from the mean, which is over the four converged seeds. $^{b}$Whisper seed~2025 exceeded walltime at $<$1 epoch (per-epoch generation-based evaluation is costly) and seed~1234 produced degenerate empty output at decode; both excluded (3/5 valid).
\end{table*}

\begin{figure*}[t]
\centering
\includegraphics[width=0.8\textwidth]{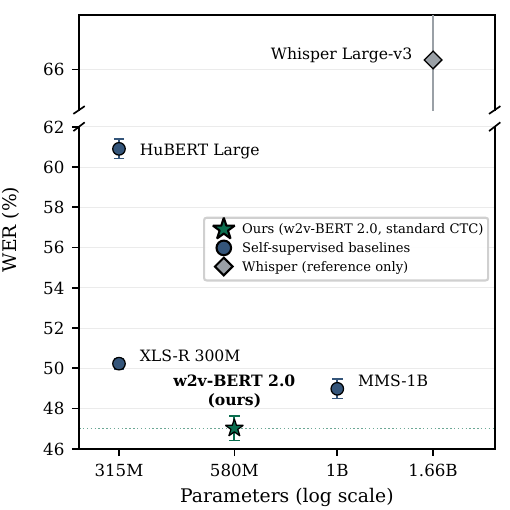}
\caption{Encoder WER vs.\ parameter count on the official Garhwali test set (five-seed means, $\pm$s.d.; broken $y$-axis). Our 580M w2v-BERT 2.0 attains the lowest WER despite smaller size; Whisper Large-v3 is reference-only (three valid seeds). Per-seed values in Table~\ref{app:tab:baseline_perseed}.}
\label{fig:scale}
\end{figure*}

\section{Per-Seed, Per-Category Error Rates}
\label{app:classerr}

We break the character errors down into phonologically defined classes to identify \emph{where} the residual errors concentrate and whether any objective shifts that distribution. Table~\ref{app:tab:ce_mean} places the three objectives side by side for both augmentation conditions; Table~\ref{app:tab:ce_perseed} gives the underlying per-seed values. Two conclusions are visible immediately. First, the error \emph{profile} is stable: in every column virama/halant is hardest ($\sim$30\%), followed by nasal and the two vowel categories ($\sim$22--26\%), with retroflex and aspirated consonants easiest ($\sim$10--16\%). Second, the objective does not reshape this profile, the matra-weighted objective leaves its own target category within 0.5 points of standard CTC, and focal CTC, far from helping, is the worst of the three on the dominant virama category and on the aggregate categorized rate. The only high-variance category is nukta, which has just 253 reference characters; we mark it $^\dagger$ and draw no conclusions from it.

\begin{table*}[t]
\centering\small\setlength{\tabcolsep}{4pt}
\caption{Per-category CER (\%), mean$\pm$s.d.\ over five seeds, all three objectives, \textbf{with} ($3\times$) and \textbf{without} speed augmentation. Substitutions and deletions only; ``Categorized'' is not corpus CER. $^\dagger$Nukta ($n{=}253$): high-variance.}
\label{app:tab:ce_mean}
\begin{tabular}{lccccccc}
\toprule
& \multicolumn{3}{c}{\textbf{With $3\times$ aug}} & & \multicolumn{3}{c}{\textbf{No aug}} \\
\cmidrule(lr){2-4}\cmidrule(lr){6-8}
\textbf{Category} & \textbf{Std} & \textbf{Focal} & \textbf{Matra} & & \textbf{Std} & \textbf{Focal} & \textbf{Matra} \\
\midrule
Matra & 22.26\,{\small$\pm$}0.20 & 22.78\,{\small$\pm$}0.55 & 22.31\,{\small$\pm$}0.77 & & 22.63\,{\small$\pm$}0.93 & 23.16\,{\small$\pm$}1.65 & 22.78\,{\small$\pm$}0.90 \\
Indep.\ vowel & 23.03\,{\small$\pm$}1.37 & 23.27\,{\small$\pm$}1.65 & 22.65\,{\small$\pm$}1.35 & & 24.32\,{\small$\pm$}0.85 & 24.41\,{\small$\pm$}1.16 & 24.01\,{\small$\pm$}1.16 \\
Virama & 29.51\,{\small$\pm$}2.59 & 32.40\,{\small$\pm$}3.59 & 29.56\,{\small$\pm$}1.95 & & 30.34\,{\small$\pm$}2.67 & 29.90\,{\small$\pm$}2.70 & 31.32\,{\small$\pm$}1.17 \\
Nasal & 25.56\,{\small$\pm$}2.46 & 26.77\,{\small$\pm$}1.27 & 26.17\,{\small$\pm$}2.12 & & 25.93\,{\small$\pm$}1.17 & 26.39\,{\small$\pm$}1.20 & 27.94\,{\small$\pm$}2.69 \\
Aspirated & 15.45\,{\small$\pm$}1.89 & 14.72\,{\small$\pm$}1.05 & 15.16\,{\small$\pm$}0.94 & & 14.74\,{\small$\pm$}0.52 & 16.08\,{\small$\pm$}1.03 & 14.41\,{\small$\pm$}1.87 \\
Retroflex & 10.71\,{\small$\pm$}0.52 & 10.83\,{\small$\pm$}0.43 & 10.47\,{\small$\pm$}0.63 & & 11.74\,{\small$\pm$}0.48 & 11.56\,{\small$\pm$}0.81 & 11.69\,{\small$\pm$}0.37 \\
Nukta$^\dagger$ & 12.88\,{\small$\pm$}6.22 & 17.63\,{\small$\pm$}8.49 & 18.73\,{\small$\pm$}6.97 & & 19.29\,{\small$\pm$}5.25 & 22.05\,{\small$\pm$}4.77 & 19.21\,{\small$\pm$}5.07 \\
\midrule
Categorized & 16.07\,{\small$\pm$}0.30 & 16.69\,{\small$\pm$}0.57 & 16.25\,{\small$\pm$}0.44 & & 16.85\,{\small$\pm$}0.47 & 17.24\,{\small$\pm$}1.26 & 16.98\,{\small$\pm$}0.63 \\
\bottomrule
\end{tabular}
\end{table*}

\noindent
For completeness, Table~\ref{app:tab:ce_perseed} gives the full per-seed values behind the mean table above, with the three objectives arranged in adjacent blocks so the seed-level stability of each category can be inspected directly.

\begin{table*}[t]
\centering\small\setlength{\tabcolsep}{2.2pt}
\caption{Per-seed per-category CER (\%), all three objectives, \textbf{with} ($3\times$) and \textbf{without} augmentation; means in Table~\ref{app:tab:ce_mean}. Seeds: 25\,=\,2025, 34\,=\,1234. $^\dagger$Nukta ($n{=}253$): high-variance.}
\label{app:tab:ce_perseed}
\begin{tabular}{l ccccc c ccccc c ccccc}
\toprule
& \multicolumn{5}{c}{\textbf{Standard CTC}} & & \multicolumn{5}{c}{\textbf{Focal CTC}} & & \multicolumn{5}{c}{\textbf{Matra-weighted}} \\
\cmidrule(lr){2-6}\cmidrule(lr){8-12}\cmidrule(lr){14-18}
\textbf{Cat.} & 42 & 123 & 777 & 25 & 34 & & 42 & 123 & 777 & 25 & 34 & & 42 & 123 & 777 & 25 & 34 \\
\midrule
\multicolumn{18}{l}{\emph{With $3\times$ speed augmentation}} \\
\midrule
Matra & 22.2 & 22.3 & 22.0 & 22.5 & 22.3 &  & 22.5 & 23.6 & 23.1 & 22.6 & 22.2 &  & 21.9 & 22.6 & 22.9 & 21.2 & 22.9 \\
Indep.\ vowel & 21.9 & 24.5 & 24.0 & 23.5 & 21.3 &  & 21.8 & 25.3 & 24.2 & 21.4 & 23.6 &  & 21.2 & 21.7 & 23.6 & 24.5 & 22.4 \\
Virama & 30.2 & 29.6 & 25.6 & 29.5 & 32.8 &  & 29.7 & 32.2 & 36.1 & 28.1 & 35.9 &  & 30.0 & 28.2 & 32.6 & 27.6 & 29.3 \\
Nasal & 23.4 & 23.2 & 25.0 & 28.1 & 28.1 &  & 25.4 & 25.9 & 26.5 & 27.5 & 28.6 &  & 22.7 & 27.5 & 26.1 & 26.3 & 28.2 \\
Aspirated & 12.7 & 17.5 & 16.7 & 14.6 & 15.8 &  & 14.1 & 16.2 & 13.5 & 14.6 & 15.3 &  & 16.0 & 14.3 & 15.2 & 14.1 & 16.2 \\
Retroflex & 10.0 & 10.5 & 11.4 & 10.6 & 11.0 &  & 10.4 & 10.9 & 11.5 & 10.9 & 10.5 &  & 10.7 & 11.4 & 10.3 & 10.4 & 9.7 \\
Nukta$^\dagger$ & 5.9 & 22.1 & 8.7 & 13.0 & 14.6 &  & 11.9 & 32.4 & 14.2 & 12.6 & 17.0 &  & 12.6 & 28.5 & 13.8 & 23.7 & 15.0 \\
\cmidrule(lr){1-18}
Categorized & 15.8 & 16.2 & 15.8 & 16.0 & 16.5 &  & 16.0 & 17.5 & 16.8 & 16.4 & 16.7 &  & 15.8 & 16.5 & 16.5 & 15.8 & 16.7 \\
\midrule
\multicolumn{18}{l}{\emph{Without augmentation}} \\
\midrule
Matra & 23.8 & 23.2 & 22.3 & 22.5 & 21.4 &  & 21.5 & 23.0 & 25.4 & 24.2 & 21.7 &  & 24.1 & 22.2 & 23.3 & 22.3 & 21.9 \\
Indep.\ vowel & 22.9 & 24.9 & 24.3 & 25.1 & 24.3 &  & 22.7 & 23.8 & 25.2 & 25.5 & 24.8 &  & 24.8 & 24.2 & 25.1 & 23.8 & 22.1 \\
Virama & 31.4 & 32.8 & 30.4 & 31.3 & 25.8 &  & 29.0 & 31.8 & 32.9 & 30.0 & 25.9 &  & 33.1 & 31.2 & 31.5 & 30.0 & 30.8 \\
Nasal & 26.8 & 26.6 & 23.9 & 26.2 & 26.1 &  & 26.3 & 25.4 & 26.0 & 28.5 & 25.8 &  & 32.4 & 26.9 & 27.4 & 27.9 & 25.1 \\
Aspirated & 15.4 & 14.2 & 14.9 & 14.9 & 14.2 &  & 16.4 & 16.1 & 17.3 & 16.2 & 14.5 &  & 14.2 & 14.9 & 17.2 & 13.6 & 12.1 \\
Retroflex & 11.4 & 12.4 & 12.1 & 11.5 & 11.3 &  & 10.8 & 11.5 & 12.9 & 11.5 & 11.2 &  & 11.7 & 11.6 & 11.1 & 11.8 & 12.1 \\
Nukta$^\dagger$ & 24.9 & 24.5 & 18.6 & 14.2 & 14.2 &  & 24.1 & 27.3 & 24.9 & 17.4 & 16.6 &  & 24.5 & 13.4 & 20.9 & 14.2 & 22.9 \\
\cmidrule(lr){1-18}
Categorized & 17.1 & 17.3 & 16.6 & 17.1 & 16.2 &  & 16.3 & 17.4 & 19.0 & 17.8 & 15.8 &  & 17.8 & 16.6 & 17.4 & 16.6 & 16.4 \\
\bottomrule
\end{tabular}
\end{table*}

\subsection{Where Does Adaptation Happen?}
\label{sec:probing}
A single-seed linear-probing analysis of the encoder layers (Appendix~\ref{app:probing}) indicates that fine-tuning primarily repurposes the \emph{upper} layers: probe error falls monotonically with depth, and the best probe moves from the middle of the base model (layer~16) to the top of the fine-tuned model (layer~24, near full-model performance), while lower layers stay close to their pretrained state. We treat this as exploratory and draw no quantitative conclusions from it.

\subsection{LLM-Assisted Cross-Model Error Categorisation}
\label{app:llm_error}

To obtain a functional view of the residual errors, grouping each reference--hypothesis mismatch by error \emph{type} rather than by character class we prompted four independent LLMs (OpenAI GPT-5.5, Claude Sonnet~4.6, Gemini~3.5 Flash, DeepSeek-V3) to categorise the standard-CTC predictions on a single seed (450 reference--hypothesis pairs; the same predictions throughout), and reconciled their outputs against the underlying pairs. We stress the limits of this analysis up front. It is \emph{single-seed}, and our central thesis is that single-seed numbers are unreliable at this corpus size; we therefore do not treat any percentage here as a stable quantity, and we draw only directional conclusions. The robustness we rely on is \emph{across models}, not across seeds: four models analysing the \emph{same} seed's predictions controls for how errors are bucketed, not for seed variance. We report it solely as a cross-check on the seed-aggregated character-class analysis of Appendix~\ref{app:classerr}, which remains our primary error analysis.

Table~\ref{app:tab:llm_error} reports one representative reconciliation (the most conservative of the four, which additionally corrected two systematic over-counts in the others). All four models independently agree on the ordering: orthographic/phonetic drift dominates ($\sim$54--60\% of error chunks across the four), acoustically-driven semantic substitution is second ($\sim$25--39\%), and word-boundary, insertion, and deletion errors are individually small. Two reconciled findings are worth isolating. (i)~\textbf{Dialect normalisation is minimal.} Although three of the four models initially placed dialect flattening at 8--12\%, reconciliation against the actual marker pairs shows only $\sim$1.4\% of error chunks ($20$ instances across $442$ content rows, in about six recurring marker pairs) are true Garhwali$\rightarrow$Hindi grammatical normalisation; the remainder are ordinary phonetic drift that lands on a Hindi-looking spelling. This supports, rather than complicates, our representational-bottleneck reading: the model is not systematically erasing dialect, it is failing on fine acoustic contrasts. (ii)~\textbf{Silence hallucination is a seed-robust, ceiling-level failure.} On the eight true-silence segments (empty reference), the system emits text in all eight cases ($100\%$), averaging tens of spurious words per segment. We verified this count directly against the predictions of all five seeds: every seed emits text on all eight silent segments ($40/40$, averaging 25--29 spurious words), so unlike the percentages above this failure is established across seeds. We report it as a rate over silent segments rather than as a share of total errors, which would understate its severity; it does not appear in our character-class profile because that profile scores only against non-empty references.

\begin{table*}[t]
\centering
\setlength{\tabcolsep}{4pt}
\caption{LLM-assisted functional error categorisation of standard-CTC predictions, single seed, reconciled across four independent models (GPT-5.5, Claude Sonnet~4.6, Gemini~3.5 Flash, DeepSeek-V3; 450 reference--hypothesis pairs; 442 content rows, 8 true-silence rows). Percentages are single-seed and reported only to convey the \emph{ordering}, which is consistent across the four models; they are not seed-aggregated and should not be read as stable quantities (see text). Silence hallucination is reported as a rate per silent segment.}
\label{app:tab:llm_error}
\begin{tabular}{lcc}
\toprule
\textbf{Error category} & \textbf{Chunks} & \textbf{\% of error chunks} \\
\midrule
Orthographic / phonetic drift      & 765 & 54.0 \\
Semantic / lexical substitution    & 555 & 39.1 \\
Word-boundary / spacing            & 33  & 2.3 \\
Deletions                          & 21  & 1.5 \\
Dialect$\rightarrow$Hindi (true)   & 20  & 1.4 \\
Insertions                         & 16  & 1.1 \\
\midrule
Silence hallucination & \multicolumn{2}{c}{$8/8$ silent segments (100\%)} \\
\bottomrule
\end{tabular}
\end{table*}

\section{Layer-wise Probing}
\label{app:probing}

\begin{figure*}[t]
\centering
\includegraphics[width=0.999\textwidth]{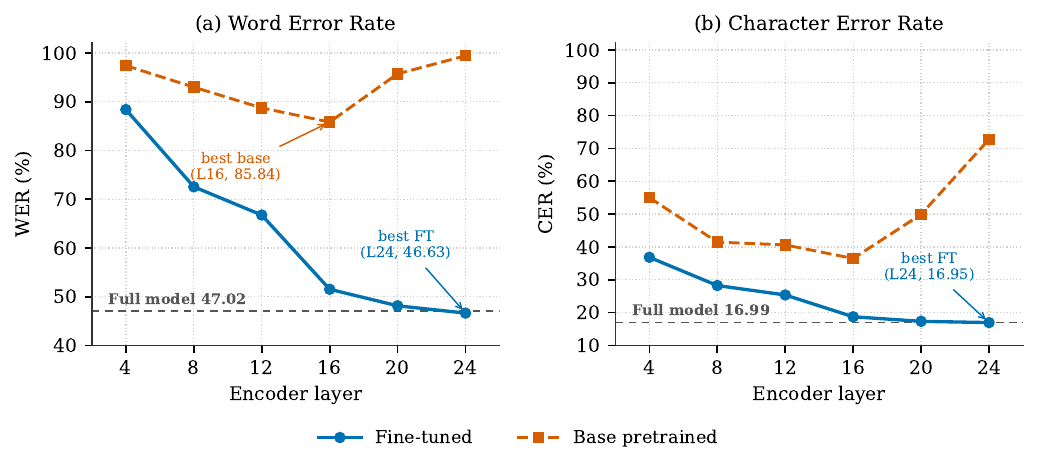}
\caption{Layer-wise linear probing of w2v-BERT 2.0 (67{,}650-parameter probe; seed 42, standard CTC). (a) WER, (b) CER for fine-tuned vs.\ base encoders; dashed line marks the five-seed mean full-model performance (this seed: 46.73/16.71). Fine-tuning repurposes the upper layers; the base model peaks mid-network. Single-seed, exploratory.}
\label{fig:probing}
\end{figure*}

To localise where fine-tuning adapts the encoder, we train a lightweight linear CTC probe (a single \texttt{Linear} layer, 67{,}650 parameters, AdamW, five epochs) on the frozen hidden states of individual encoder layers, for both the base pretrained w2v-BERT 2.0 and the fine-tuned standard-CTC model (seed 42). Figure~\ref{fig:probing} reports probe WER and
CER at layers 4--24.

Two patterns emerge. In the \emph{base} model, decodable linguistic information is highest in the middle of the network (best probe at layer~16, 85.8\% WER) and degrades sharply in the final layers (99.5\% at layer~24) the characteristic ``autoencoder'' profile of self-supervised speech encoders, whose top layers specialise toward the pretraining objective rather than toward phonetic content. Fine-tuning reshapes exactly this region: probe error decreases monotonically with depth and the best probe moves to the \emph{top} layer (layer~24, 46.6\% WER, essentially full-model performance: 46.73\% for this seed, 47.0\% five-seed mean). The improvement from fine-tuning grows with depth, negligible at layer~4 ($-9$ WER points) but large at layers~20 and~24 ($-48$ and $-53$ points) indicating that adaptation is concentrated in the upper encoder layers while the lower layers remain close to their pretrained state. As a single-seed analysis this is exploratory and we read it qualitatively, but the direction is consistent with the standard view that task fine-tuning primarily repurposes the upper layers of SSL speech encoders.

\section{Objective Definitions}
\label{app:objectives_math}

For completeness we give the full form of the two non-standard objectives summarised in Section~\ref{sec:objectives}. The length-normalised, temperature-scaled confidence used by Focal CTC is
\begin{equation}
  p \;=\; \exp\!\left(-\,\frac{\ell / L}{\tau}\right),
  \label{eq:focal_p}
\end{equation}
where $\ell$ is the per-utterance CTC loss, $L$ the target length (in tokens), and $\tau$ the temperature; this $p$ enters the focal objective of Eq.~\eqref{eq:focal_loss}. The matra-weighted objective adds a class-weighted auxiliary term to the focal loss,
\begin{equation}
  \mathcal{L} \;=\; \mathcal{L}_{\text{base}}
                 \;+\; \lambda\,\mathcal{L}_{\text{aux}},
  \label{eq:matra_total}
\end{equation}
with $\mathcal{L}_{\text{base}}$ the focal CTC loss of Eq.~\eqref{eq:focal_loss} and
\begin{equation}
  \mathcal{L}_{\text{aux}}
    \;=\; -\,\frac{\sum_{f} w_{c(f)}\,\log p\!\left(t_f \mid f\right)}
                  {\sum_{f} w_{c(f)}},
  \label{eq:matra_aux}
\end{equation}
the sum over non-blank frames $f$, with $t_f$ the emitted token and $c(f)$ its phonological class.

\section{Reproducibility Details}
\label{app:repro}

Table~\ref{app:tab:config} lists the configuration shared by all primary w2v-BERT 2.0 systems; only the training objective and the presence of augmentation differ across the runs compared in the paper, so any difference in results is attributable to those two factors rather than to optimisation settings. 

\paragraph{Compute.}
All primary systems were trained on a single NVIDIA A100-SXM4 (80\,GB) with 4 CPU cores and 48\,GB of system RAM. Under early stopping (patience 5 on validation WER), runs converged in roughly 12--13 epochs. Per-seed wall-clock training time was $\sim$1.4\,h (median; range 1.3--1.6\,h) without augmentation and $\sim$2.9\,h (median; range 2.3--5.6\,h) with $3\times$ speed augmentation about a $2\times$ increase in wall-clock despite the training set tripling to 14{,}334 utterances, since throughput (20--29 training utterances/s) and the early-stopping epoch count absorb part of the nominal data increase. Validation over the 666 utterances took $\sim$9\,s per epoch ($\sim$75 utterances/s). Inference cost is identical across all objectives: every system decodes with the same single-pass greedy CTC head, so the objective, augmentation, and transfer comparisons in the paper differ only in training, not in decoding cost.

\begin{table*}[t]
\centering\setlength{\tabcolsep}{4pt}
\caption{Configuration shared by all primary (w2v-BERT 2.0) systems. Only the objective and augmentation setting vary across compared runs.}
\label{app:tab:config}
\begin{tabular}{ll}
\toprule
\textbf{Setting} & \textbf{Value} \\
\midrule
Encoder & w2v-BERT 2.0 (\texttt{facebook/w2v-bert-2.0}), 580M \\
Encoder layers & 24, all fine-tuned \\
Frozen & feature-projection layer only \\
Features & 80-dim log-Mel (SeamlessM4T), 16\,kHz \\
Vocabulary & 66 tokens (63 chars + word-delim, unk, pad) \\
Splits (official) & 4{,}778 / 666 / 450 (train/val/test) \\
Speed aug & 0.9$\times$,1.0$\times$,1.1$\times$ $\rightarrow$ 14{,}334 train \\
Encoder LR & $3\times10^{-5}$ \\
Head LR & $10^{-3}$ \\
Schedule & 10\% linear warmup, then linear decay \\
Optimizer & AdamW \\
Batch (effective) & 32 \\
Precision & BF16 \\
Max epochs & 20, early stop (patience 5) on val WER \\
Decoding & per-utterance greedy \\
Seeds & 42, 123, 777, 2025, 1234 \\
Hardware & 1$\times$ NVIDIA A100-SXM4 (80GB), 4 CPU, 48\,GB RAM \\
\bottomrule
\end{tabular}
\end{table*}

\noindent
Table~\ref{app:tab:sig} reports the pairwise seed-level tests in full. All three Holm-corrected $p$-values exceed 0.05; standard vs.\ focal sits exactly at the five-seed floor of $0.1875$ (Section~\ref{sec:protocol}) because standard CTC wins on all five seeds, so the closest call again points away from focal being beneficial, and the substantive evidence is the paired differences and effect sizes of Table~\ref{tab:power}.

\paragraph{Early-stopping metric.}
A methodological difference from \citet{dhasmana-etal-2026-dialect} is the early-stopping criterion: we stop on validation WER, they on validation CER, a metric increasingly argued to be more appropriate for multilingual and Indic ASR \citep{thennal2025advocating}. To confirm this does not drive our numbers, we re-run standard CTC under CER-based stopping over the same five seeds and official splits, obtaining 47.56$\pm$0.74 WER, within noise of our WER-stopped 47.02$\pm$0.61 (paired Wilcoxon $p=0.81$). The early-stopping metric therefore does not materially change the outcome.

\begin{table*}[t]
\centering\setlength{\tabcolsep}{4pt}
\caption{Pairwise significance for the augmented objective comparison. Primary test: per-seed paired Wilcoxon signed-rank on corpus WER, Holm--Bonferroni corrected. No pair is significant at $\alpha=0.05$.}
\label{app:tab:sig}
\begin{tabular}{lcc}
\toprule
\textbf{Pair} & \textbf{$\Delta$WER} & \textbf{$p$ (Holm)} \\
\midrule
Standard vs.\ Focal     & $+0.81$ & $0.19$ \\
Standard vs.\ Matra     & $+0.40$ & $0.38$ \\
Focal vs.\ Matra        & $-0.42$ & $0.38$ \\
\bottomrule
\end{tabular}
\end{table*}

\subsection{Per-Utterance Bootstrap Cross-Check}
\label{app:bootstrap}

As a secondary check on the objective comparison, we bootstrap the \emph{per-utterance} WER. Pooling the per-utterance WERs of all five seeds (2{,}210 scored utterances per objective, after excluding utterances with empty reference text, matching our corpus-WER scoring), we resample utterances with replacement (2{,}000 replicates) and take the 2.5/97.5 percentiles (Table~\ref{app:tab:bootstrap}). We stress that this is a \emph{secondary} analysis: a per-utterance bootstrap measures only test-set sampling noise and ignores seed-to-seed variation, so it understates the true uncertainty of the comparison and its intervals are correspondingly narrow. The seed-level paired tests of Section~\ref{sec:objective_comparison} remain our primary evidence. Even so, the cross-check agrees with the primary analysis in ordering: the intervals overlap substantially and match the corpus-level means (standard $<$ matra $<$ focal). A paired Wilcoxon on the same pooled per-utterance WERs does flag standard vs.\ focal (Holm $p=0.008$); we report this for completeness only, since pooling treats the same utterance under different seeds as independent and overstates significance, and its direction again favours standard CTC. Note that macro-averaged per-utterance WER runs about two points above the corpus (micro-averaged) WER of Table~\ref{tab:objective_multiseed}, as short utterances with a single error carry disproportionate weight.

\begin{table*}[t]
\centering\setlength{\tabcolsep}{6pt}
\caption{Per-utterance bootstrap WER (\%) pooled over five seeds: mean and 95\%
percentile interval (2{,}000 resamples). Secondary cross-check only; intervals
understate uncertainty because they exclude seed variation. Ordering and overlap
agree with the seed-level null of Section~\ref{sec:objective_comparison}.}
\label{app:tab:bootstrap}
\begin{tabular}{lcc}
\toprule
\textbf{Objective} & \textbf{Per-utt WER} & \textbf{95\% CI} \\
\midrule
Standard CTC          & 49.31 & [48.39, 50.31] \\
Focal CTC             & 50.31 & [49.38, 51.28] \\
Matra-weighted (ours) & 49.74 & [48.81, 50.74] \\
\bottomrule
\end{tabular}
\end{table*}

\begin{table*}[t]
\centering
\setlength{\tabcolsep}{6pt}
\renewcommand{\arraystretch}{1.35}
\caption{Example outputs of the standard-CTC system (seed~42) on the official Garhwali test set, sampled \emph{by error level}: for each of seven per-utterance CER targets (0--50\%) we show the test utterance whose CER is nearest that target, among references of 12--45 characters. Examples are selected purely by the per-utterance error rate, not hand-picked by error type or quality, and therefore span the full range from exact matches to heavily degraded output. REF is the reference transcript, HYP the greedy decode.}
\label{app:tab:examples}
\begin{tabular}{r l p{0.74\textwidth}}
\toprule
\textbf{CER} & & \textbf{Transcript} \\
\midrule
\multirow{2}{*}{0\%}  & REF & \dev{यख तीन आदमी दिखेण लग्यां} \\
                      & HYP & \dev{यख तीन आदमी दिखेण लग्यां} \\
\midrule
\multirow{2}{*}{5\%}  & REF & \dev{अपणी अगल बगल दुकानी मा कखी भी मिली जांदी} \\
                      & HYP & \dev{अपड़ी अगल बगल दुकानी मा कखी भी मिली जांदी} \\
\midrule
\multirow{2}{*}{10\%} & REF & \dev{यख मकान फर गेट लगयुं} \\
                      & HYP & \dev{यख मकान पर गेट लग्युं} \\
\midrule
\multirow{2}{*}{15\%} & REF & \dev{व्यायाम सी लोग स्वस्थ रंदा} \\
                      & HYP & \dev{बयाम सी लोग स्वस्थ रंदा} \\
\midrule
\multirow{2}{*}{20\%} & REF & \dev{यख छकी बेंच लगीं फांडु वाइट बोर्ड तख चिपकायूँ} \\
                      & HYP & \dev{यख छक्की बेंच लगीं फांड वाइट बोर्ड तख छककयुं} \\
\midrule
\multirow{2}{*}{30\%} & REF & \dev{लोग वख पाणी ल्योण थै भरी गांगजी जोंद थैई} \\
                      & HYP & \dev{लोग वा पाणी लन थै भरी गंगा जी लन थैई} \\
\midrule
\multirow{2}{*}{50\%} & REF & \dev{उंठा कुछ नई तख देखिनी का} \\
                      & HYP & \dev{औट कुछ न ते दिखणकद} \\
\bottomrule
\end{tabular}
\end{table*}

\end{document}